\documentclass[10pt,twocolumn]{article}

\usepackage{microtype}
\usepackage{graphicx}
\usepackage{booktabs} 

\usepackage{hyperref}
\usepackage{amsmath,amssymb}
\usepackage{enumitem}
\usepackage{subcaption}

\usepackage[colorinlistoftodos]{todonotes}

\title{GAP: Generalizable Approximate Graph Partitioning Framework}
\author
{
Azade Nazi$^{\dag}$\footnote{Members of the Google Brain Residency Program} ,
Will Hang$^{\dag\dag}$,
Anna Goldie$^{\dag}$,
Sujith Ravi$^{\dag}$, 
Azalia Mirhoseini$^{\dag}$
\\
$^{\dag}$Google Research;
$^{\dag\dag}$Stanford University
\\
$^{\dag}$\small{\{azade,~agoldie,~sravi,~azalia\}@google.com},
$^{\dag\dag}$\small{\{willhang\}@stanford.edu}
}
\date{}

\begin{document}
\maketitle
\begin{abstract}



Graph partitioning is the problem of dividing the nodes of a graph into balanced partitions while minimizing the edge cut across the partitions. Due to its combinatorial nature, many approximate solutions have been developed, including variants of multi-level methods and spectral clustering. We propose GAP, a {\it Generalizable Approximate Partitioning} framework that takes a deep learning approach to graph partitioning. We define a differentiable loss function that represents the partitioning objective and use backpropagation to optimize the network parameters. Unlike baselines that redo the optimization per graph, GAP is capable of generalization, allowing us to train models that produce performant partitions at inference time, even on unseen graphs. Furthermore, because we learn the representation of the graph while jointly optimizing for the partitioning loss function, GAP can be easily tuned for a variety of graph structures. We evaluate the performance of GAP on graphs of varying sizes and structures, including graphs of widely used machine learning models (e.g., ResNet, VGG, and Inception-V3), scale-free graphs, and random graphs. We show that GAP achieves competitive partitions while being up to 100 times faster than the baseline and generalizes to unseen graphs.
\end{abstract}

\section{Introduction}
\label{sec:intro}


Graph partitioning is an important optimization problem with numerous applications in domains spanning computer vision, VLSI design, biology, social networks, transportation networks and more. The objective is to find balanced partitions of a graph while minimizing the number of edge cut. This problem is NP-complete which is formulated as a discrete optimization problem and solutions are generally derived using heuristics and approximation algorithms. Some notable approaches include multi-level methods and spectral partitioning methods~\cite{karypis_1998, karypis_1999, karypis_2000, miettinen_2006, andersen2006, chung2007gp}.

In this work, we introduce a learning based approach, GAP, for the continuous relaxation of the problem. We define a differentiable loss function which captures the objective of partitioning a graph into disjoint balanced partitions while minimizing the number of edge cut across those partitions. We train a deep model to optimize for this loss function. The optimization is done in an unsupervised manner without the need for labeled datasets.

Our approach, GAP, does not assume anything about the graph structure (e.g., sparse vs. dense, or scale-free). Instead, GAP learns and adapts to the graph structure using graph embedding techniques while optimizing the partitioning loss function. This representation learning allows our approach to be self-adaptive without the need for us to design different strategies for different types of graphs.

Our learning based approach is also capable of generalization, meaning that we can train a model on a set of graphs and then use it at inference time on unseen graphs of varying sizes. In particular, we show that when GAP is trained on smaller graphs (e.g., 1k nodes), it transfers what it learned to much larger ones (e.g, 20k nodes). This generalization allows trained GAP models to quickly infer partitions on large unseen graphs, whereas baseline methods have to redo the entire optimization for each new graph. 

In summary, this paper makes the following contributions:
\begin{itemize}[leftmargin=*]
    \item We propose GAP, a {\it Generalizable Approximate Partitioning} framework, which is an unsupervised learning approach to the classic problem of balanced graph partitioning. We define a differentiable loss function for partitioning that uses a continuous relaxation of the normalized cut. We then train a deep model and apply backpropagation to optimize the loss.
    \item GAP models can produce efficient partitions on unseen graphs at inference time. Generalization is an advantage over existing approaches which must redo the entire optimization for each new graph.
    \item GAP leverages graph embedding techniques \cite{kipf2017semi, HamiltonYL17} and learns to partition graphs based on their underlying structure, allowing it to generate efficient partitions across a wide variety of graphs.
    \item To encourage reproducible research, we provide source code in the supplementary materials and are in the process of open-sourcing the framework.
    \item We show that GAP achieves competitive partitions while being up to 100 times faster than top performing baselines on a variety of synthetic and real-world graphs with up to 27000 nodes. 
\end{itemize}

\section{Related Work}
\label{sec:relWork}
\noindent\textbf{Graph Partitioning}:
Graph partitioning is an important combinatorial optimization problem that has been exhaustively studied. The most widely used graph partitioning algorithms generate partitions by performing operations on the input graph until convergence~\cite{andersen2006, chung2007gp}. On the other hand, multilevel partitioning approaches first reduce the size of the graph by collapsing nodes and edges, then partition on the smaller graph, and finally expand the graph to recover the partitioning for the original graph~\cite{karypis_2000, karypis_1999, karypis_1998, miettinen_2006}. These algorithms are shown to provide high-quality partitions~\cite{miettinen_2006}.

Another approach is to use simulated annealing. \cite{van_1990} proposed mean field annealing, which combines simulated annealing with Hopfield neural networks. \cite{kawamoto2018mean} studied a different formulation of graph partitioning in which a graph is generated by a statistical model, and the task is to infer the preassigned group labels of the generative model. They developed a mean-field theory of a minimal graph neural network architecture for this version of the problem.

This line of inquiry formulates graph partitioning as a discrete optimization problem, while our GAP framework is one of the first deep learning approaches for the continuous relaxation of the problem. Moreover, GAP generalizes to unseen graphs, generating partitions on the fly, rather than having to redo the optimization per graph.

\noindent\textbf{Clustering}:
Given a set of points, the goal of clustering is to identify groups of similar points. Clustering problems with different objectives such as self-balanced k-means and balanced min-cut have been exhaustively studied~\cite{liu_2017, chen_2017, chang_2014}. One of the most effective techniques for clustering is spectral clustering, which first generates node embeddings in the eigenspace of the graph Laplacian, and then applies k-means clustering to these vectors \cite{Shi_2000, ng2002spectral, von2007tutorial}.


However, generalizing clustering to unseen nodes and graphs is nontrivial. To address generalization, SpectralNet \cite{shaham2018spectralnet} is a deep learning approach to spectral clustering which generates spectral embeddings for unseen data points. Other deep learning approaches for clustering attempt to encode the input in a way that is amenable to clustering by k-means or Gaussian Mixture Models~\cite{yang2017, xie2016unsupervised, zheng2016variational, dilokthanakul2016deep}.

Although related, graph clustering and graph partitioning are different problems in that graph clustering attempts to maximize locality of clusters, whereas graph partitioning seeks to preserve locality while maintaining balance among partitions. Our approach also treats the partitioning problem as an end-to-end learning problem with a differentiable loss, whereas the aforementioned approaches generate embeddings that are then clustered using non-differentiable techniques like k-means.

\noindent\textbf{Device Placement}:
The practical significance of graph partitioning is demonstrated by the task of device placement for TensorFlow computation graphs, where the objective is to minimize execution time by assigning operations to devices. 
\cite{azalia_2017} proposed a reinforcement learning method to optimize device placement for TensorFlow graphs. They used a seq2seq policy to assign operations to devices.
The execution time of the generated placements is then used as a reward signal to optimize the policy. A hierarchical model for device placement has been proposed in \cite{azalia_2018}, where the graph partition and placement are learned jointly. While this work also uses a neural network to learn the partitions, their objective is to optimize the runtime of the resulting partitions, forcing them to use policy gradient to optimize their non-differentiable loss function.
\section{Problem Definition and Background}
\label{sec:probDef}
Let $G=(V,E)$ be a graph where $V = \{v_i\}$ and $E = \{e(v_i, v_j) | v_i \in V, v_j \in V\}$ are the set of nodes and edges in the graph. Let $n$ be the number of nodes. A graph $G$ can be partitioned into $g$ disjoint sets $S_1, S_2, \dots S_g$, where the union of the nodes in those sets are $V$ ($\bigcup_{k=1}^{g} S_k = V$), and each node belongs to only one set ($\bigcap_{k=1}^{g} S_k = \emptyset$), by simply removing edges connecting those sets. 

\noindent\textbf{Minimum Cut:}
The total number of edges that are removed from $G$ in order to form disjoint sets is called \emph{cut}. Given sets $S_k$, and $\bar{S_k}$, the $\emph{cut}(S_k, \bar{S_k})$ is formally defined as:

\begin{equation} \label{equ:cut}
\small
\emph{cut}(S_k, \bar{S_k}) = \sum_{v_i\in S_k, v_j\in \bar{S_k}} e(v_i,v_j)
\end{equation}
This formula can be generalized to multiple disjoint sets $S_1, S_2, \dots S_g$, where $\bar{S_k}$ is the union of all sets except $S_k$.

\begin{equation} \label{equ:cut_multiple}
\small
 \emph{cut}(S_1, S_2, \dots S_g) = \frac{1}{2}\sum_{i=k}^{g} \emph{cut}(S_k, \bar{S_k})
\end{equation}

\noindent\textbf{Normalized Cut:}
The optimal partitioning of a graph that minimizes the cut (Equation~\ref{equ:cut_multiple}) is a well-studied problem and there exist efficient polynomial algorithms for solving it~\cite{Papadimitriou_1982}. However, the minimum cut criteria favors cutting nodes whose degree are small and leads to unbalanced sets/partitions.
To avoid such bias, normalized cut ($\emph{Ncut}$), which is based on the graph conductance, has been studied by~\cite{Shi_2000, ZhangR18}, where the cost of a cut is computed as a fraction of the total edge connections to all nodes.

\begin{equation} \label{equ:Ncut}
\small
 \emph{Ncut}(S_1, S_2, \dots S_g) = \sum_{k=1}^{g} \frac{\emph{cut}(S_k, \bar{S_k})}{\emph{vol}(S_k, V)}
\end{equation}
Where $\emph{vol}(S_k, V) = \sum_{v_i\in S_k, v_j\in V} e(v_i,v_j)$, i.e., total degree of nodes belong to $S_k$ in graph $G$. 

One way to minimize the normalized cut is based on the eigenvectors of the graph Laplacian which has been studied in~\cite{Shi_2000, ZhangR18}. Previous research has shown that across a wide range of social and information networks, the clusters with the smallest graph conductance are often small~\cite{LeskovecLDM09, ZhangR18}. Regularized spectral clustering has been proposed by~\cite{ZhangR18} to address this problem.

In this paper, however, we propose GAP as an unsupervised learning approach with a differentiable loss function that can be trained to find balanced partitions with minimum normalized cuts. We show that GAP enables generalization to unseen graphs. 

\section{Generalizable Approximate Partitioning}
\label{sec:tech}

\begin{figure*}[h]
\centering
\includegraphics[scale=0.65]{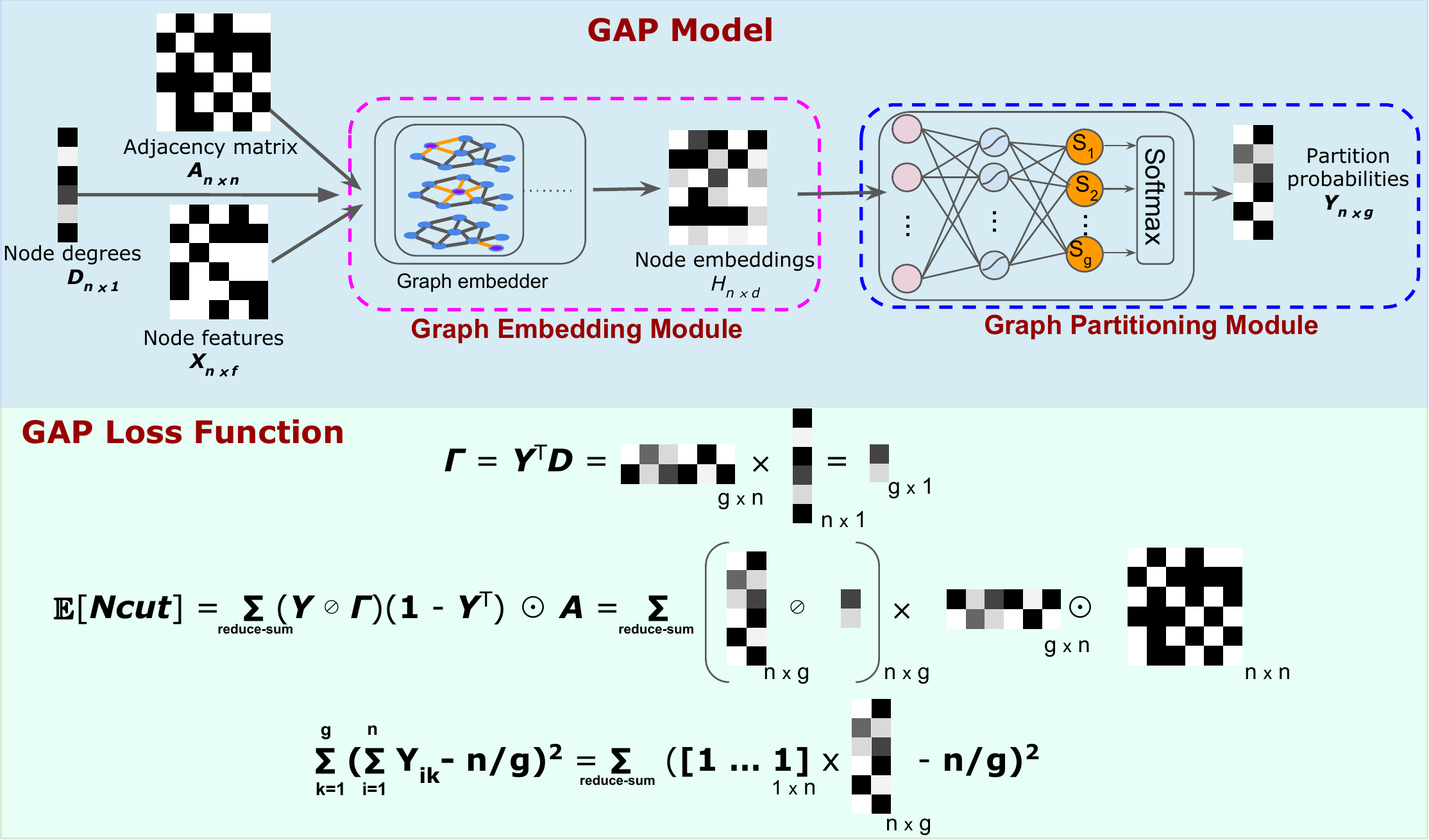}
\caption{Generalizable Approximate graph Partitioning (GAP) Framework (see Section~\ref{sec:tech} for more details).}
\label{fig:framework}
\end{figure*}

We now introduce the \emph{Generalizable Approximate Partitioning} framework (GAP). As shown in Figure \ref{fig:framework}, GAP has two main components: graph representation learning for generating partition probabilities per node (the model), and a differentiable formulation of the normalized cut objective (the loss function). GAP enables us to train a neural network to optimize a previously undifferentiable objective by generating balanced partitions with minimum edge-cut. We first present the loss function before discussing the model.

\subsection{GAP Loss Function}
\label{sec:loss}
We assume that our model returns $Y \in \mathbb{R}^{n \times g}$ where $Y_{ik}$ represents the probability that node $v_i \in V$ belongs to partition $S_k$. We propose a loss function based on $Y$ to calculate the normalized cut in Equation~\ref{equ:Ncut} and evaluate the balancedness of the partitions. Later in subsection~\ref{sec:framework}, we discuss the model that generates $Y$. 

\noindent\textbf{Normalized Cut:} 
As we discussed in Section~\ref{sec:probDef}, $\emph{cut}(S_k, \bar{S}_k)$ is the number of edges $e(v_i, v_j)$, where $v_i \in S_k$ and $v_j \notin S_k$. Let $Y_{ik}$ be the probability that node $v_i$ belongs to partition $S_k$. The probability that node $v_j$ does not belong to partition $S_k$ would be $1 - Y_{jk}$. Therefore, $\mathbb{E}[\emph{cut}(S_k, \bar{S_k})]$ can be formulated by Equation~\ref{equ:Ycut1}, where $\mathcal{N}(v_i)$ is the set of nodes adjacent to $v_i$ (visual illustration in Figure~\ref{fig:framework}).

\begin{equation}
\label{equ:Ycut1}
\small
\begin{aligned}
\mathbb{E}[\emph{cut}(S_k, \bar{S}_k)] &= \sum_{\substack{v_i \in S_k \\ v_j \in \mathcal{N}(v_i)}} \sum_{z=1}^{g}  Y_{iz} (1-Y_{jz})
\end{aligned}
\end{equation}

Since the set of adjacent nodes for a given node can be retrieved from the adjacency matrix of graph $A$, we can rewrite Equation~\ref{equ:Ycut1} as follows:

\begin{equation}
\label{equ:Ycut}
\small
\begin{aligned}
\mathbb{E}[\emph{cut}(S_k, \bar{S}_k)] &= \sum_\text{reduce-sum} Y_{:,k} (1-Y_{:,k})^\intercal \odot A
\end{aligned}
\end{equation}

The element-wise product with the adjacency matrix $(\odot \ A)$ ensures that only the adjacent nodes are considered. Moreover, the result of $Y_{:,k} (1-Y_{:,k})^\intercal \odot A$ is an $n \times n$ matrix and $\mathbb{E}[\emph{cut}(S_k, \bar{S}_k)]$ is the sum over all of its elements.

From Equation~\ref{equ:Ncut}, $\emph{vol}(S_k, V)$ is the sum over the degree of all nodes that belong to $S_k$. Let $D$ be a column vector of size $n$ where $D_i$ is the degree of the node $v_i \in V$. Given $Y$, we can calculate the $\mathbb{E}[\emph{vol}(S_k, V)]$ as follows:

\begin{equation}
\label{equ:assoc}
\small
\begin{aligned}
\Gamma = Y^\intercal D \\
\mathbb{E}[\emph{vol}(S_k, V)] &= \Gamma_k
\end{aligned}
\end{equation}

Where $\Gamma$ is a vector in $\mathbb{R}^g$, and $g$ is the number of partitions. 

With $\mathbb{E}[\emph{cut}(S_k, \bar{S}_k)]$ and $\mathbb{E}[\emph{vol}(S_k, V)]$ from Equations~\ref{equ:Ycut} and~\ref{equ:assoc}, we can calculate the expected normalized cut in Equation~\ref{equ:Ncut} as follows:

\begin{equation}
\label{equ:YNcut}
\small
\begin{aligned}
\mathbb{E}[\emph{Ncut}(S_1, S_2, \dots S_g)] &= \sum_\text{reduce-sum} (Y \oslash \Gamma) (1-Y)^\intercal \odot A
\end{aligned}
\end{equation}

$\oslash$ is element-wise division and the result of $(Y \oslash \Gamma) (1-Y)^\intercal \odot A$ is an $n \times n$ matrix where $\mathbb{E}[\emph{cut}(S_1, S_2, \dots S_g)]$ is the sum over all of its elements. 

\noindent\textbf{Balanced Cut:} So far, we have shown how one can calculate the expected normalized cut of a graph given the matrix $Y$ (probabilities of nodes belonging to partitions). Here, we show that given $Y$ we can also
evaluate how balanced those partitions are.

Given the number of nodes in the graph $|V| = n$ and the number of partitions $g$, to have balanced partitions the number of nodes per partition should be $\frac{n}{g}$. The sum of the columns in $Y$ gives us the expected number of nodes in each partition due to the fact that $Y_{ik}$ represents the probability that node $v_i \in V$ belongs to partition $S_k$. Thus, for the balanced partitions we minimize the following error:

\begin{equation}
\label{equ:bal}
\small
\begin{aligned}
\sum_{k=1}^{g} (\sum_{i=1}^{n} Y_{ik} - \frac{n}{g})^2 = \sum_\text{reduce-sum} (\mathbf{1}^\intercal Y - \frac{n}{g})^2
\end{aligned}
\end{equation}

Combining expected normalized cut (Equation~\ref{equ:YNcut}) with the balanced partition error (Equation~\ref{equ:bal}), we have the following loss function:

\begin{equation}
\label{equ:loss}
\small
\begin{aligned}
\mathcal{L} = \sum_\text{reduce-sum} (Y \oslash \Gamma) (1-Y)^\intercal \odot A + \sum_\text{reduce-sum} (\mathbf{1}^\intercal Y - \frac{n}{g})^2
\end{aligned}
\end{equation}

Next, we discuss the GAP neural model that finds the graph partition $Y$ to minimize the loss in Equation~\ref{equ:loss}.

\subsection{The GAP Model}
\label{sec:framework}

The GAP model ingests a graph definition, generates node embeddings that leverage local graph structure, and projects each embedding into logits that define a probability distribution to minimize the expected normalized cut (Equation~\ref{equ:loss}). 

\noindent\textbf{Graph Embedding Module}:
The purpose of the graph embedding module is to learn node embeddings using the graph structure and node features. Recently, there have been several advances on applying graph neural networks for node embedding and classification tasks using approaches such as Graph Convolution Network~\cite{kipf2017semi} (GCN), GraphSAGE~\cite{HamiltonYL17}, Neural Graph Machines~\cite{NGL17}, Graph Attention Networks~\cite{GAT2018} and other variants. In this work, we leverage GCN and GraphSAGE to learn graph representations across a variety of graphs, which helps with generalization. 

\noindent\emph{GCN}:
~\cite{kipf2017semi} showed that untrained GCN with random weights can serve as a powerful feature extractor for graph nodes. In our implementation, we used a 3-layer GCN with weight matrices ($\mathbf{W}^{(l)}$) using Xavier initialization described in~\cite{Xavier_2010}.

\begin{equation*}
\label{equ:gcn}
\small
\begin{aligned}
Z = \text{tanh}(\hat{A} \text{~tanh}(\hat{A} \text{~tanh}(\hat{A}X\mathbf{W}^{(0)}) \mathbf{W}^{(1)} )  \mathbf{W}^{(2)} )
\end{aligned}
\end{equation*}
where $\hat{A} = \tilde{D}^{-\frac{1}{2}} \tilde{A} \tilde{D}^{-\frac{1}{2}}$, $\tilde{A} = A + I_{n}$ is the adjacency matrix of the undirected graph $G$ with added self-connections. $I_{n}$ is the identity matrix, and $\tilde{D}_{i,i} = \sum_j \tilde{A}_{ij}$. The input feature matrix $X$ depends on the graph. In TensorFlow computation graphs, each operation type (such as MatMul, Conv2d, Sum, etc.) would be a feature.

\noindent\emph{GraphSAGE}:
~\cite{HamiltonYL17} developed a node embedding technique that generates high dimensional graph node representations based on node input features. Central to this technique is \emph{sample and aggregate}, where given a node $v_i$, we sample a set of $v_i$'s neighbors from $\mathcal{N}(v_i)$, and aggregate their representations (with max pooling) to generate an embedding for the sampled neighbors of $v_i$. This neighbor representation, along with the representation of $v_i$ itself, is combined to generate a new representation for $v_i$. Iterating this process multiple times results in message passing among nodes for an increasing number of hops. 

Our implementation of GraphSAGE is based on Algorithm 1 in ~\cite{HamiltonYL17}. For each message passing step $k$, we perform the following operations per node $v_i \in V$:

$$\mathbf{h}^k_{\mathcal{N}(v_i)} = \text{maxpool}(\{\mathbf{W}^{\text{agg}}_k\mathbf{h}_{v_j}^{k - 1} + \mathbf{b}^{\text{agg}}_k , \forall v_j \in \mathcal{N}(v_i)\})$$
$$\mathbf{h}_{v_i}^k = \text{relu}(\mathbf{W}_k^{\text{proj}}[\mathbf{h}_{v_i}^{k - 1}, \mathbf{h}_{\mathcal{N}(v_i)}^k] + \mathbf{b}^{\text{agg}}_k)$$
$$\mathbf{h}^k_{v_i} = \mathbf{h}^k_{v_i} / ||\mathbf{h}^k_{v_i}||_2$$

where agg and proj denote the aggregation and projection matrices respectively.

\noindent\textbf{Graph Partitioning Module}:
The second module in our GAP framework is responsible for partitioning the graph, taking in node embeddings and generating the probability that each node belongs to partitions $S_1, S_2, ..., S_g$ (Y in Figure~\ref{fig:framework}). This module is a fully connected layer followed by softmax, trained to minimize Equation~\ref{equ:loss}. 

We also note that for particularly large graphs, it is possible to optimize on randomly sampled minibatches of nodes from the larger graph. Furthermore, it is possible to stop gradient flow from the partitioning module to the embedding module, resulting in unsupervised node embeddings.
\begin{table*}[h]
\centering
\renewcommand{\arraystretch}{1.0}
\begin{tabular}{|p{2.0cm}||p{1.8cm}|p{2.2cm}||p{1.8cm}|p{2.2cm}|} 
\hline
\multicolumn{1}{ |c|| }{\textbf{Computation graphs}} & \multicolumn{2}{ c|| }{\textbf{hMETIS}} & \multicolumn{2}{ c| }{\textbf{GAP}}\\
\hline \centering
{\sf Name} & {\sf Edge cut } & {\sf Balancedness} & {\sf Edge cut } & {\sf Balancedness} \\
\hline
\hline \centering
{\sf \textit{VGG}} & {\sf 0.05 } & {\sf 0.99} & {\sf  0.04} & {\sf 0.99}\\
\hline \centering
{\sf  \textit{MNIST-conv}} & {\sf 0.05} & {\sf 0.99} & {\sf 0.05} & {\sf  0.99}\\
\hline \centering
{\sf  \textit{ResNet}} & {\sf 0.04} & {\sf 0.99} & {\sf 0.04} & {\sf  0.99}\\
\hline \centering
{\sf  \textit{AlexNet}} & {\sf 0.05} & {\sf 0.99} & {\sf 0.05} & {\sf  0.99}\\
\hline \centering
{\sf  \textit{Inception-v3}} & {\sf 0.04} & {\sf 0.99} & {\sf 0.04} & {\sf  0.99}\\
\hline
\end{tabular}
\caption{Performance of GAP against hMETIS. Number of partitions is three and we run hMETIS and GAP over a given graph.For edge cut lower is better, for balancedness higher is better.}
\label{tbl:perf}
\end{table*}

\begin{figure*}[h]
    \centering
    \begin{subfigure}{0.23\textwidth}
        \includegraphics[width=40mm,height=26mm]{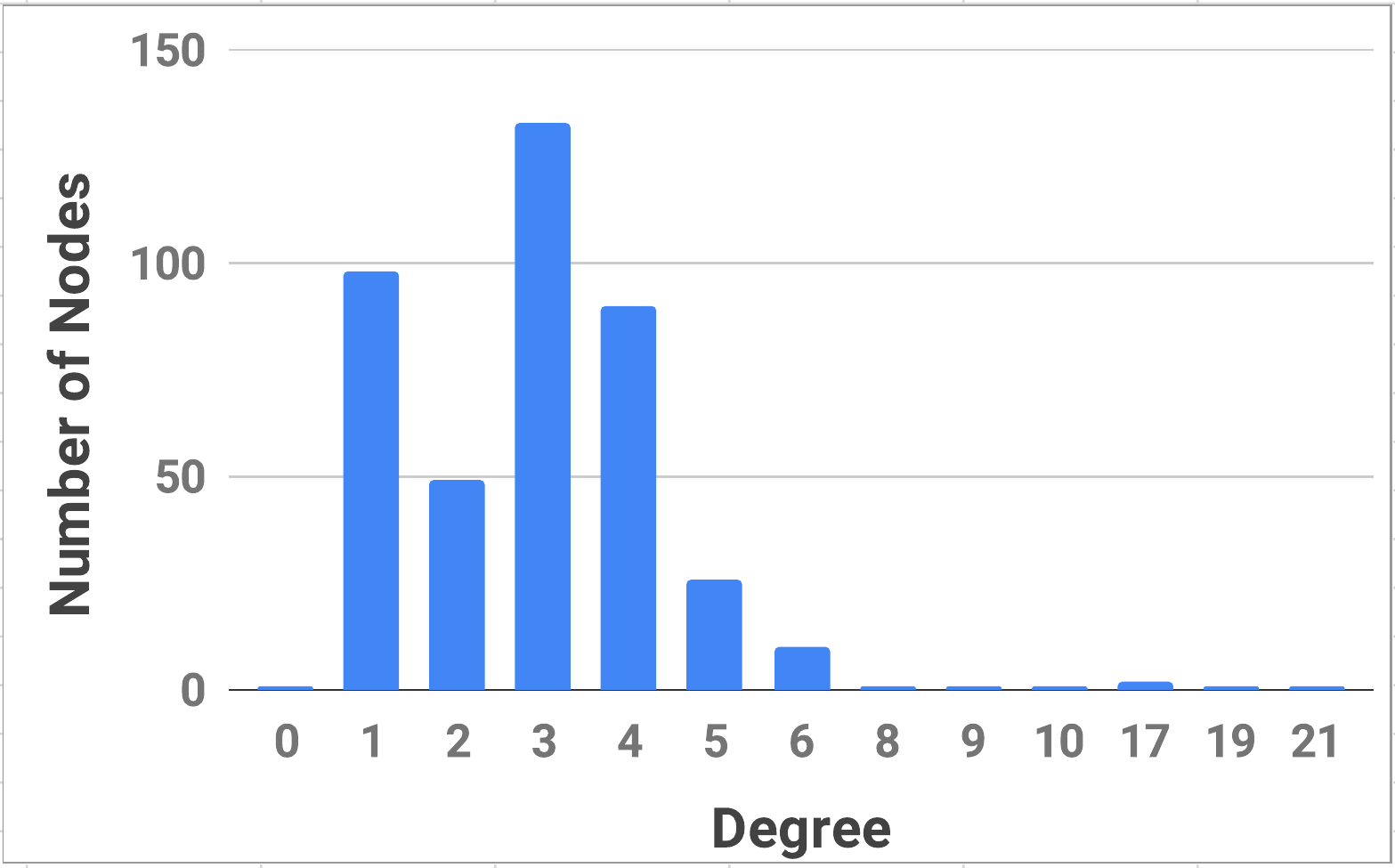}
        \caption{\small{MNIST-conv}}
  \label{fig:mnist-dist}
    \end{subfigure}
    \hspace{0.5mm}
    \begin{subfigure}{0.23\textwidth}
        \includegraphics[width=40mm,height=26mm]{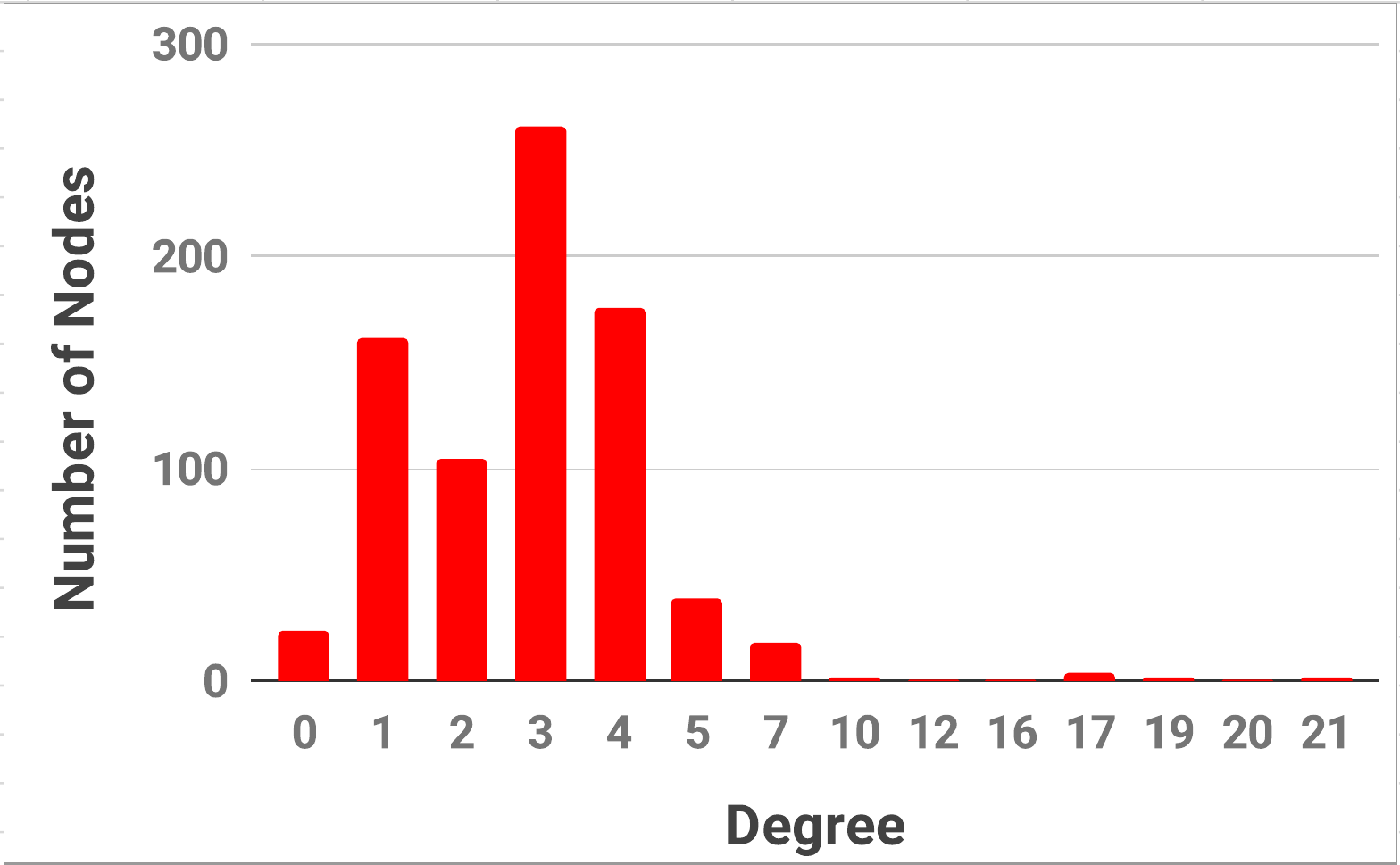}
        \caption{\small{AlexNet}}
        \label{fig:alexnet-dist}
    \end{subfigure}
    \hspace{0.5mm}
    \begin{subfigure}{0.23\textwidth}
        \includegraphics[width=40mm,height=26mm]{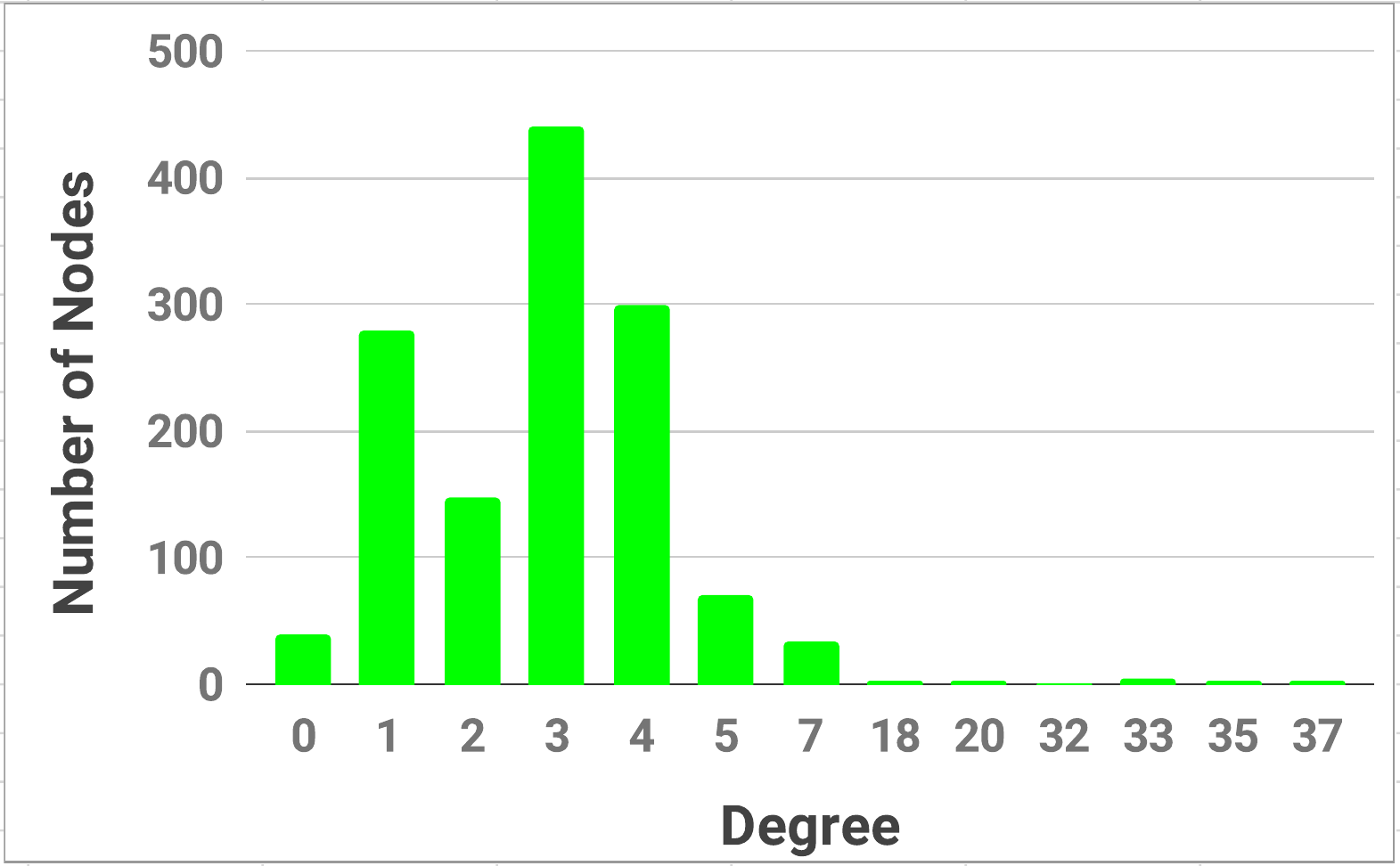}
        \caption{\small{VGG}}
        \label{fig:vgg-dist}
    \end{subfigure}
    \hspace{0.5mm}
    \begin{subfigure}{0.23\textwidth}
        \includegraphics[width=40mm,height=26mm]{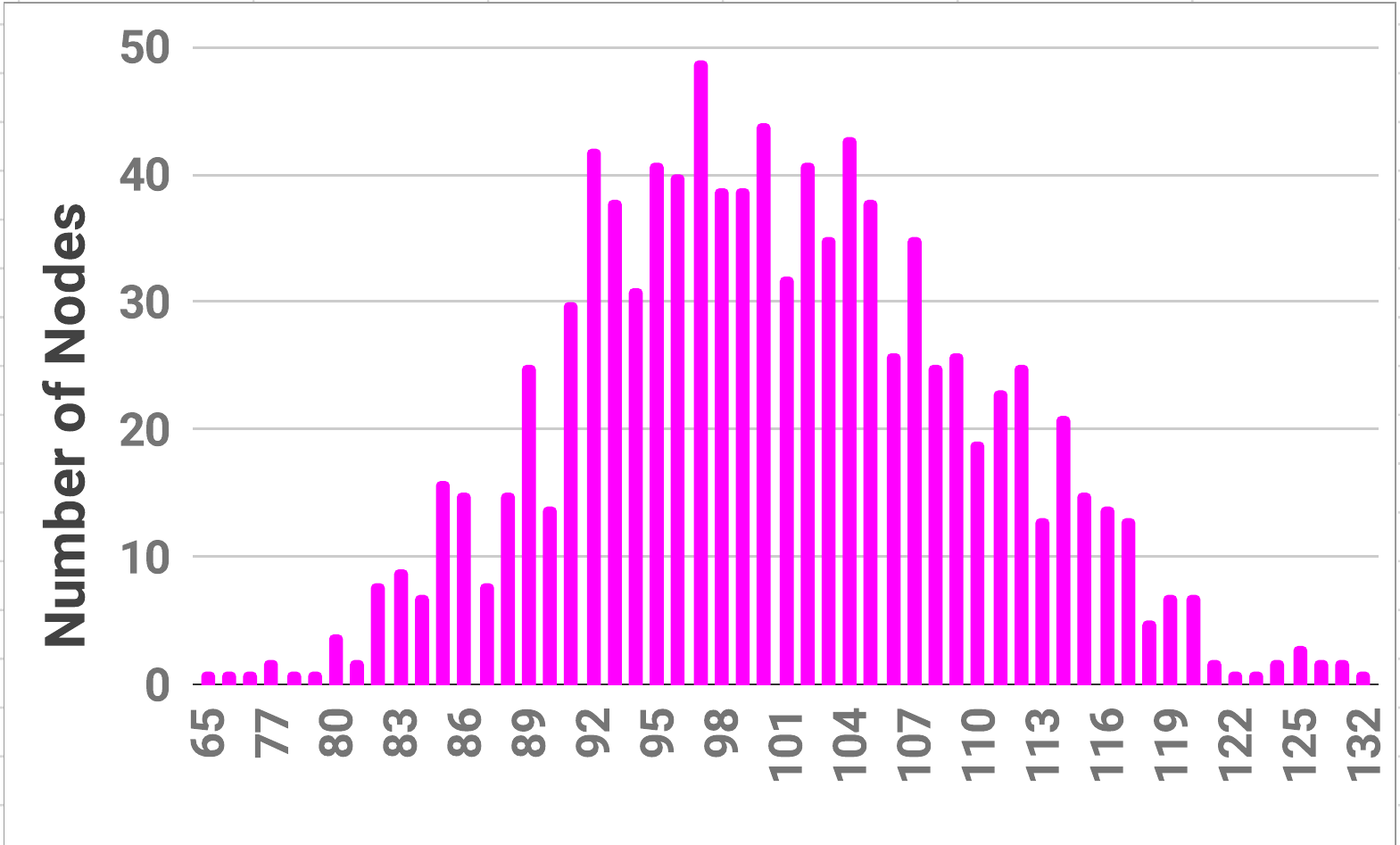}
        \caption{\small{Random graph}}
        \label{fig:random-1000-dist}
    \end{subfigure}
    \caption{Degree histogram of \textit{MNIST-conv}, \textit{VGG}, \textit{AlexNet} and synthetic \textit{Random} graphs. \textit{Random} graphs are denser than the others.}
    \label{fig:deg-hist}
\end{figure*}

\begin{figure}[h]
\centering
\includegraphics[scale=0.44]{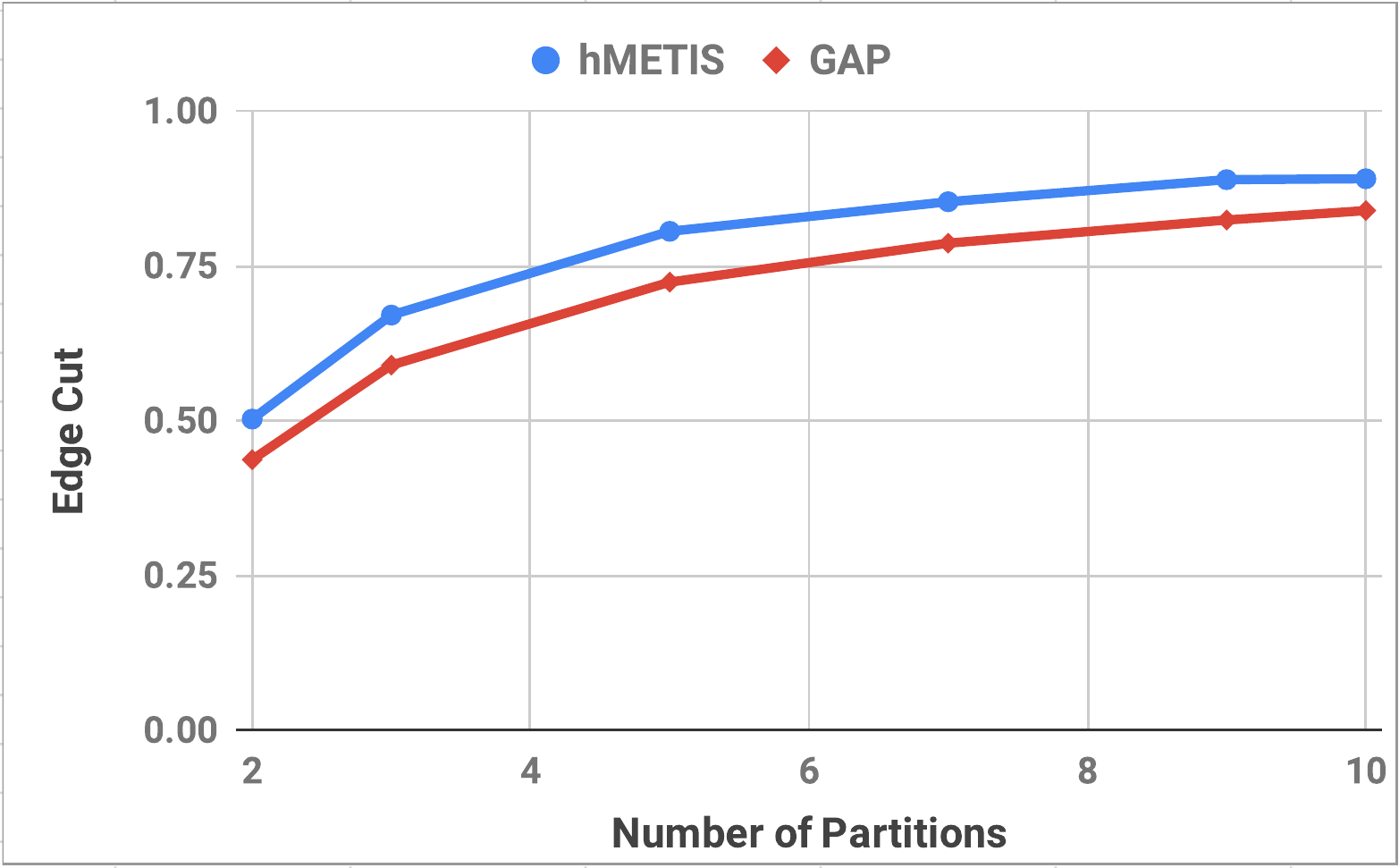}
\caption{Edge cut of the partitions on random graphs by varying the number of partitions using GAP and hMETIS. Both GAP and hMETIS produce 99\% balanced partitions.}
\label{fig:edge_cut_random}
\end{figure}

\section{Experiments}
\label{sec:exp}
The main goals of our experiments are to (a) evaluate the performance of the GAP framework against hMETIS~\cite{karypis_2000}, a widely used partitioner that uses multilevel partitioning and (b) evaluate
the generalizability of GAP over unseen graphs and provide insights on how the structural similarities between train and test graphs affect the generalization performance. Source code is provided for reproducibility and is in the process of being open-sourced.

\subsection{Setup}
\label{sec:setup}
We conducted experiments on real and synthetic graphs. Specifically, we use five widely used TensorFlow graphs. We also generate \textit{Random} as well as \textit{Scale-free} graphs as synthetic datasets to show the effectivenesss of GAP on graphs with different structures.

\noindent\textbf{Real Datasets}
\begin{itemize}[leftmargin=*]
\item{ \textit{ResNet}~\cite{resnet_2016} is a deep convolutional network with residual connections to avoid vanishing gradients. The TensorFlow implementation of $\emph{ResNet\_v1\_50}$ with 50 layers contains $20,586$ operations.}
\item{\textit{Inception-v3}~\cite{inception_2017} consists of multiple blocks, each composed of several convolutional and pooling layers. The TensorFlow graph of this model contains $27,114$ operations.}
\end{itemize}
\begin{itemize}[leftmargin=*]
\item{\textit{AlexNet}~\cite{alexnet_2012} consists of 5 convolutional layers, some of which are followed by max-pooling layers, and 3 fully-connected layers with a final softmax. The TensorFlow graph of this model has $798$ operations.}
\item \textit{MNIST-conv} has 3 convolutional layers for the $\emph{MNIST}$ classification task. The TensorFlow graph of this model contains $414$ operations.
\item{\textit{VGG}~\cite{vgg_2014} contains 16 convolutional layers. The TensorFlow graph of $\emph{VGG}$ contains $1,325$ operations.}
\end{itemize}

\noindent\textbf{Synthetic Datasets} 
\begin{itemize}[leftmargin=*]
\item{ \textit{Random}: Randomly generated networks of size $10^3$ and $10^4$ nodes using the $\emph{Erd{\"o}s--R{\'e}nyi}$ model~\cite{erdos_1960}, where the probability of having an edge between any two nodes is $0.1$.}
\item{ \textit{Scale-free}: Randomly generated scale-free networks of size $10^3$ and $10^4$ nodes using NetworkX~\cite{hagberg2008} (A scale-free network is a network whose degree distribution follows a power law~\cite{Bollobas2003}).}
\end{itemize}

\begin{table*}[h]
\centering
\renewcommand{\arraystretch}{1.0}
\begin{tabular}{|p{1.4cm}|p{1.8cm}||p{1.4cm}|p{2.0cm}||p{1.4cm}|p{2.0cm}||p{1.4cm}|p{2.0cm}|} 
\hline
\multicolumn{2}{ |c|| }{\textbf{Computation graphs}} & \multicolumn{2}{ c|| }{\textbf{\textit{AlexNet}}} & \multicolumn{2}{ c|| }{\textbf{\textit{Inception-v3}}} & \multicolumn{2}{ c| }{\textbf{\textit{ResNet}}}\\
\hline \centering
{\sf Name} & {\sf Embedding} & {\sf Edge cut } & {\sf Balancedness} & {\sf Edge cut } & {\sf Balancedness} & {\sf Edge cut } & {\sf Balancedness}\\
\hline
\hline \centering
{\sf  \textit{GAP-op}} & {\sf  \textit{-}} & {\sf 0.16} & {\sf 0.71} & {\sf 0.24} & {\sf  0.74} & {\sf  0.45} & {\sf 0.90}\\
\hline \centering
{\sf  \textit{GAP-id}} & {\sf  \textit{GCN  offline}} & {\sf 0.28} & {\sf 0.97} & {\sf 0.19} & {\sf  0.98} & {\sf  0.17} & {\sf 0.93}\\
\hline \centering
{\sf  \textit{GAP-op}} & {\sf  \textit{GCN  offline}} & {\sf 0.07} & {\sf \textbf{0.99}} & {\sf 0.12} & {\sf  0.98} & {\sf  0.11} & {\sf 0.94}\\
\hline \centering
{\sf  \textit{GAP-op}} & {\sf  \textit{GraphSAGE offline}} & {\sf 0.07} & {\sf \textbf{0.99}} & {\sf 0.08} & {\sf \textbf{0.99}} & {\sf  0.09} & {\sf 0.95}\\
\hline \centering
{\sf  \textit{GAP-op}} & {\sf  \textit{GraphSAGE trained}} & {\sf \textbf{0.06}} & {\sf \textbf{0.99}} & {\sf \textbf{0.06}} & {\sf \textbf{0.99}} & {\sf  \textbf{0.08}} & {\sf \textbf{0.98}}\\
\hline
\end{tabular}
\caption{Generalization results: GAP is trained on \textit{VGG} and validated on \textit{MNIST-conv}. During inference, the model is applied to unseen TensorFlow graphs: \textit{ResNet}. \textit{Inception-v3}, and \textit{AlexNet}. In \textit{GAP-id}, we use node index features, while in \textit{GAP-op}, we use TensorFlow operation types as features. According to Table~\ref{tbl:perf}, the ground truth for \textit{VGG}, \textit{MNIST-conv}, and \textit{AlexNet} is $99\%$ balanced partitions with $5\%$ edge cut and for \textit{ResNet} and \textit{Inception-v3}, it is $99\%$ balanced partitions with $4\%$ edge cut. \textit{GAP-op} with \textit{GraphSAGE trained} (last row) generalizes better than the other models.}
\label{tbl:gen}
\end{table*}

\begin{figure*}[h]
    \centering
    \begin{subfigure}{0.45\textwidth}
        \includegraphics[width=75mm,height=20mm]{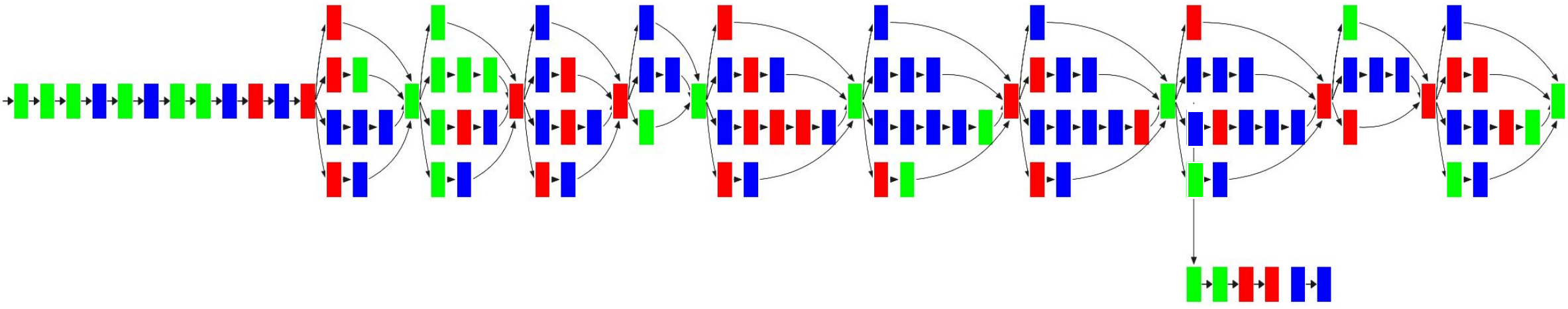}
        \caption{\small{Training a GAP model on \textit{Inception-v3}, and testing on the same computation graph (\textit{Inception-v3}) achieves 99\% balanced partitions with 4\% edge cut (Table~\ref{tbl:perf})}}
        \label{fig:inception}
    \end{subfigure}
    \hspace{1mm}
    \begin{subfigure}{0.45\textwidth}
        \includegraphics[width=75mm,height=20mm]{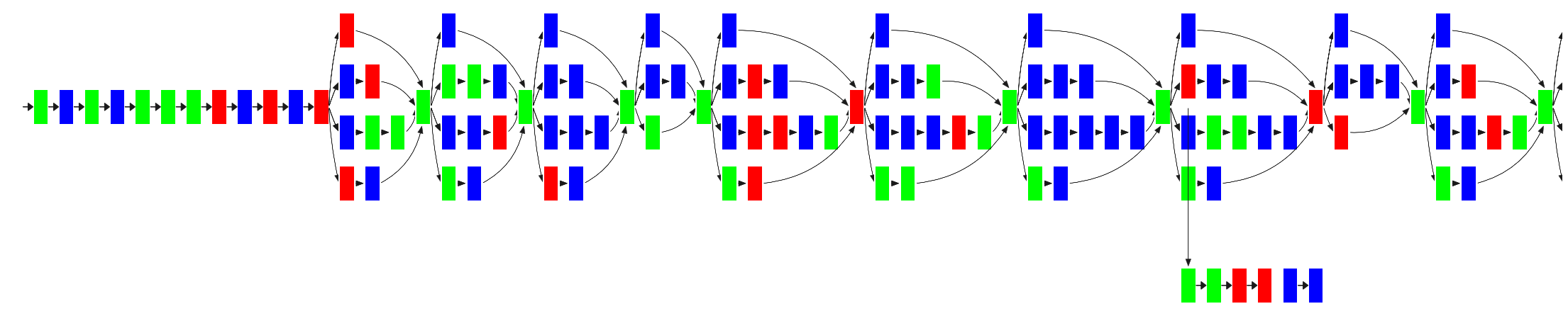}
        \caption{\small{Generalization: training a GAP model on \textit{VGG}, and testing it on unseen graphs (\textit{Inception-v3}) achieves 99\% balanced partitions with 6\% edge cut (last row of Table~\ref{tbl:gen})}}
        \label{fig:inception-gen}
    \end{subfigure}
    \caption{GAP partitioning of the \textit{Inception-v3} (a) using the trained model on \textit{Inception-v3} and (b) the trained model on \textit{VGG}. Number of partitions is three and they are denoted by colors. We only show the nodes whose operation type is convolution. }
    \label{fig:inception_gen_vis}
\end{figure*}

\begin{figure}[h]
\centering
\includegraphics[scale=0.43]{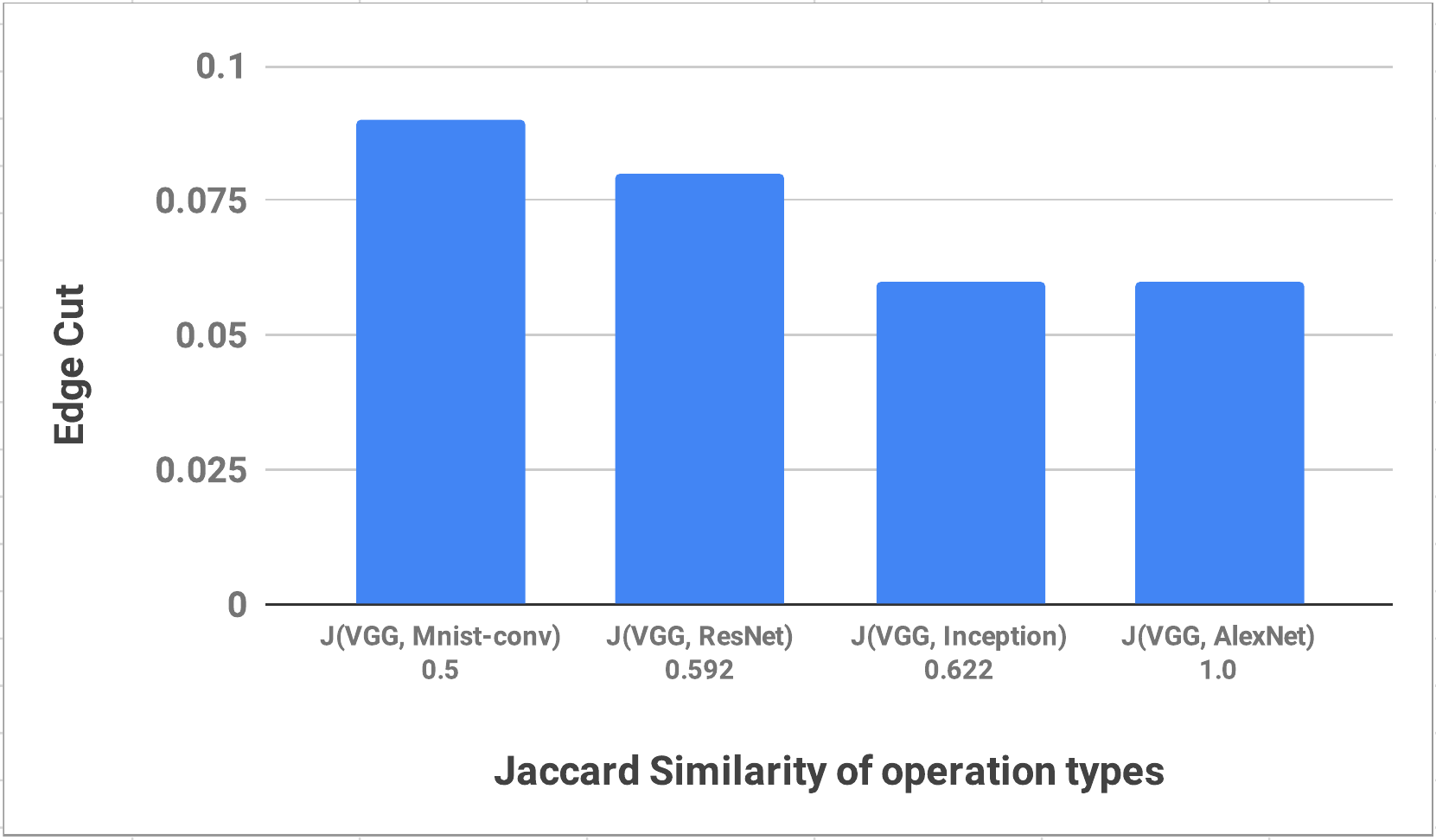}
\caption{Here, GAP is trained on \textit{VGG} and is tested on other computation graphs. We observe that the Jaccard similarity between the operation types in \textit{VGG} and other graphs affects the generalization of GAP. Higher Jaccard similarities between the train and validation/test datasets enable GAP to find better partitions with smaller edge cut.}
\label{fig:similarity_op}
\end{figure}

\begin{figure*}[h]
    \centering
    \begin{subfigure}{0.31\textwidth}
        \includegraphics[scale=0.3]{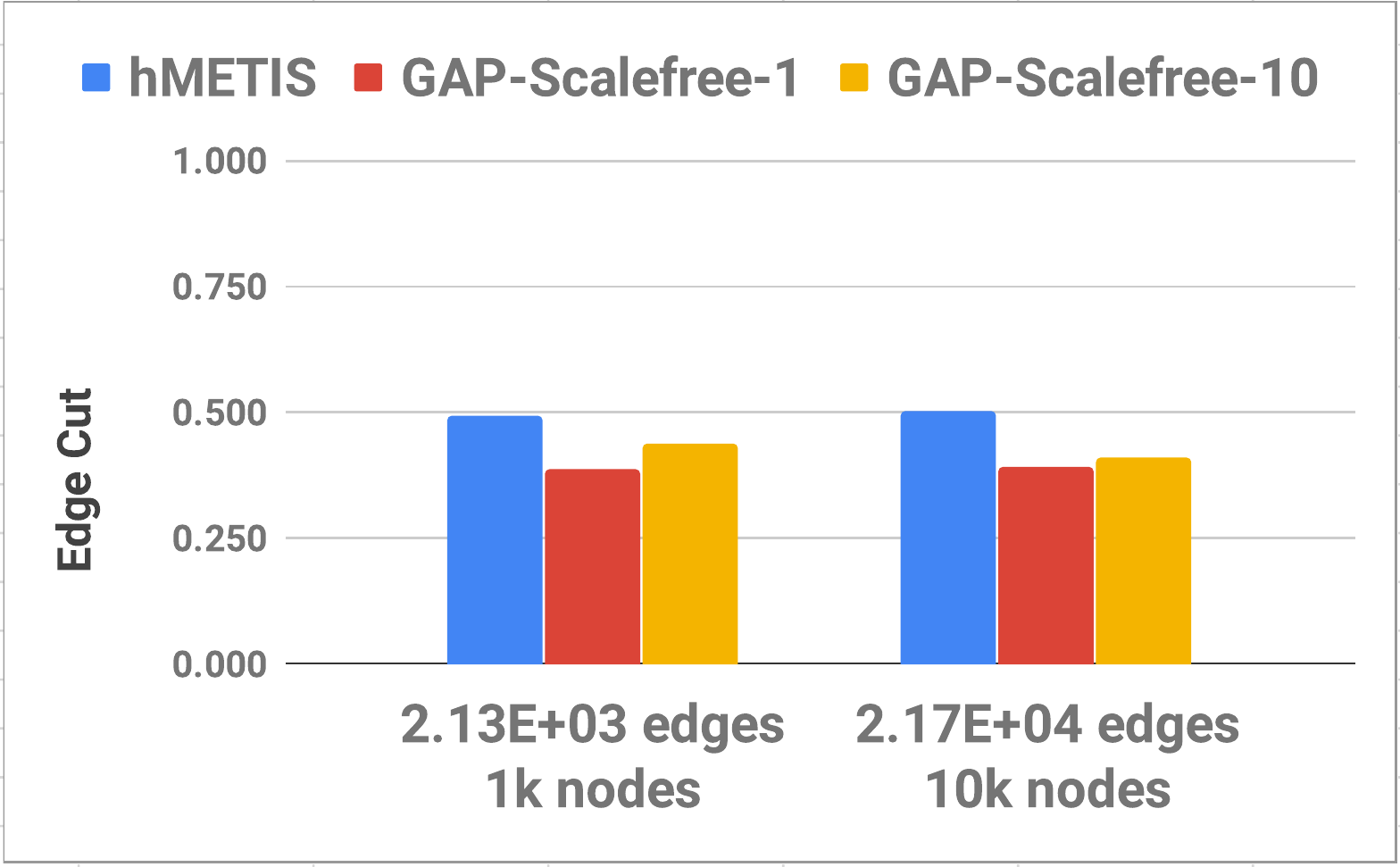}
        \caption{\small{Edge cut of GAP vs hMETIS. GAP partitions unseen graphs of 1k and 10k nodes with smaller edge cut.}}
        \label{fig:scalefree-edgecut}
    \end{subfigure}
    \hspace{1mm}
    \begin{subfigure}{0.33\textwidth}
        \includegraphics[scale=0.3]{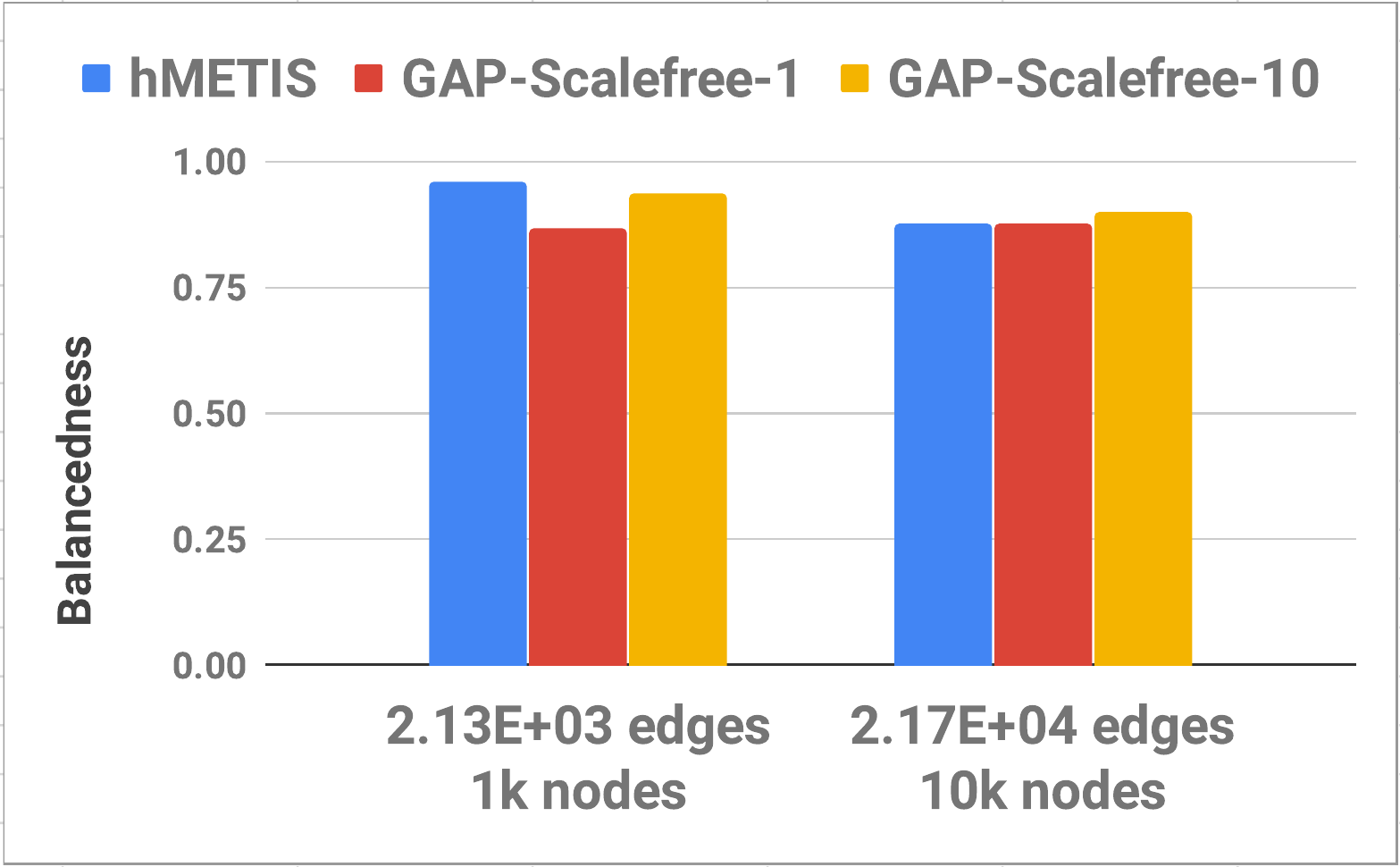}
        \caption{\small{Balancedness of GAP vs hMETIS. GAP-Scalefree-10 (trained on more graphs) improves the balancedness.}}
        \label{fig:scalefree-bal}
    \end{subfigure}
    \hspace{1mm}
    \begin{subfigure}{0.3\textwidth}
        \includegraphics[scale=0.3]{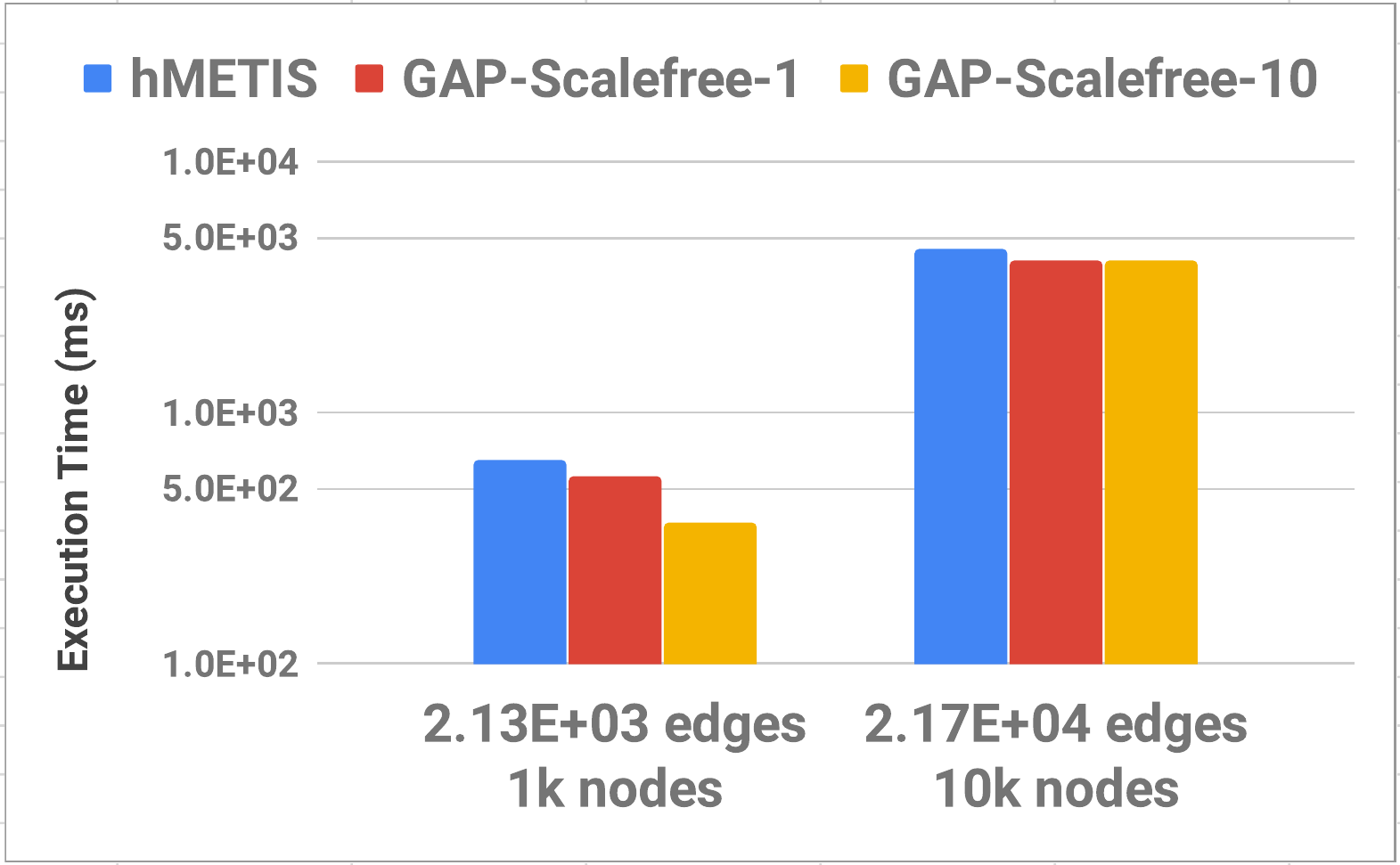}
        \caption{\small{Inference time of GAP vs running time of hMETIS. GAP is slightly faster than hMETIS.}}
        \label{fig:scalefree-exetime}
    \end{subfigure}
    \caption{Generalization of GAP on scale-free graphs. GAP-Scalefree-1 is trained on only one scale-free graph, while GAP-Scalefree-10 is trained on 10 scale-free graphs. The result is the average over the 5 scale-free graphs of 1k and 10k nodes. GAP-Scalefree-10 is slightly faster than hMETIS and it produces partitions which are as balanced as hMETIS partitions but with smaller edge cut.}
    \label{fig:scale-free}. 
\end{figure*}

\begin{figure*}[h]
    \centering
    \begin{subfigure}{0.3\textwidth}
        \includegraphics[scale=0.3]{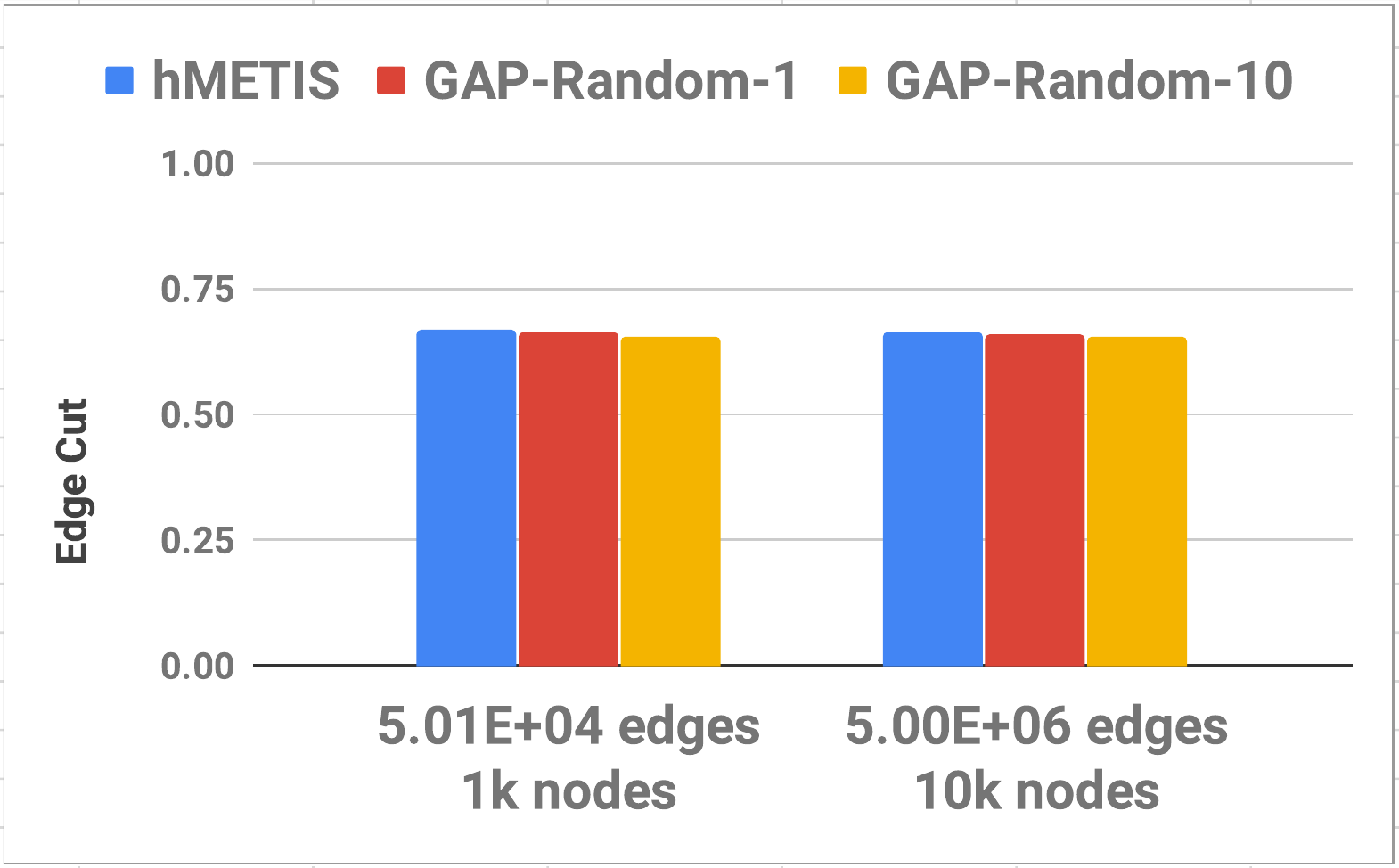}
        \caption{\small{The edge cut of GAP is slightly lower than that of hMETIS.}}
        \label{fig:random-edgecut}
    \end{subfigure}
    \hspace{1mm}
    \begin{subfigure}{0.3\textwidth}
        \includegraphics[scale=0.3]{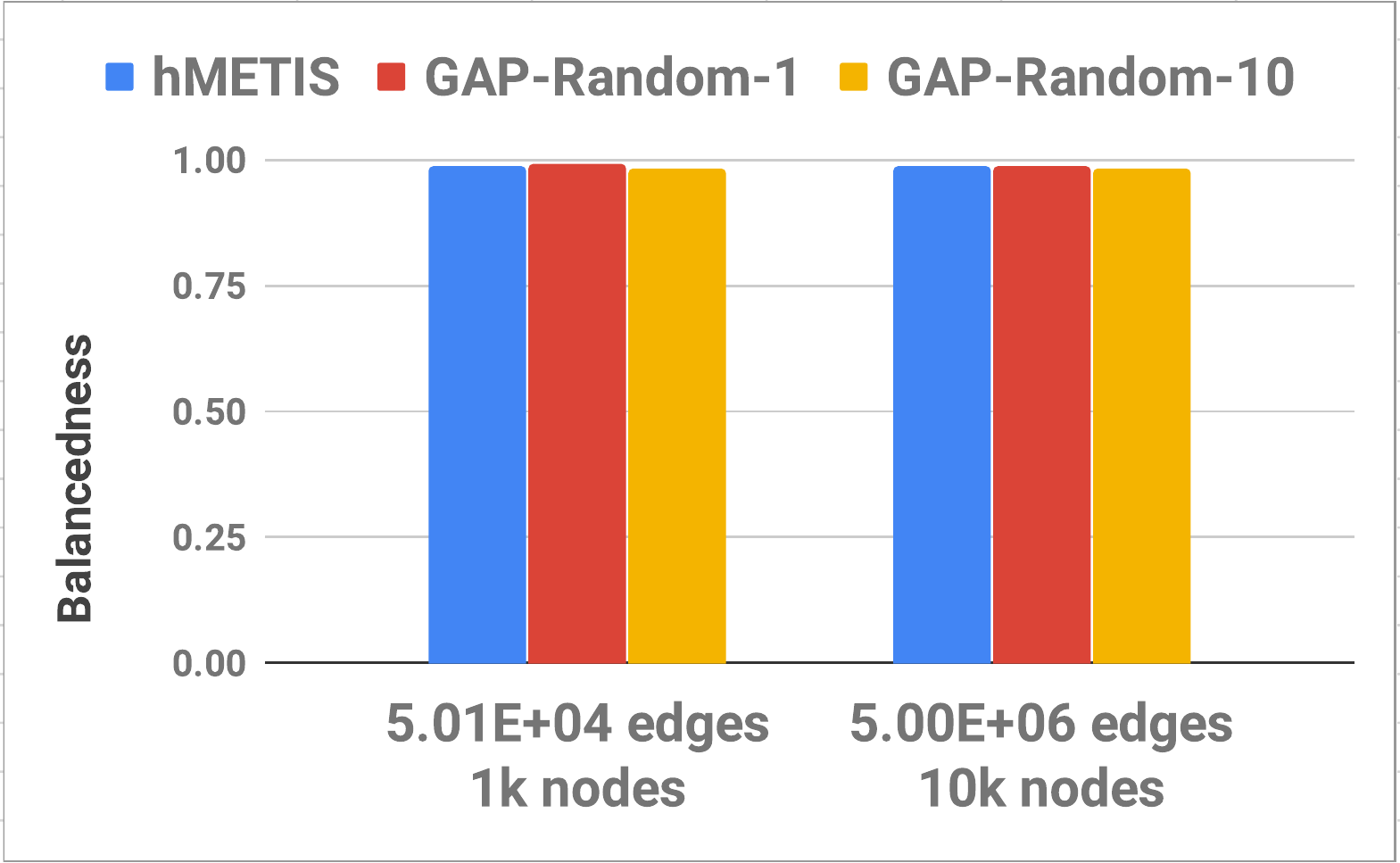}
        \caption{\small{The Balancedness of GAP and hMETIS are almost equal (99\%).}}
        \label{fig:random-bal}
    \end{subfigure}
    \hspace{1mm}
    \begin{subfigure}{0.3\textwidth}
        \includegraphics[scale=0.3]{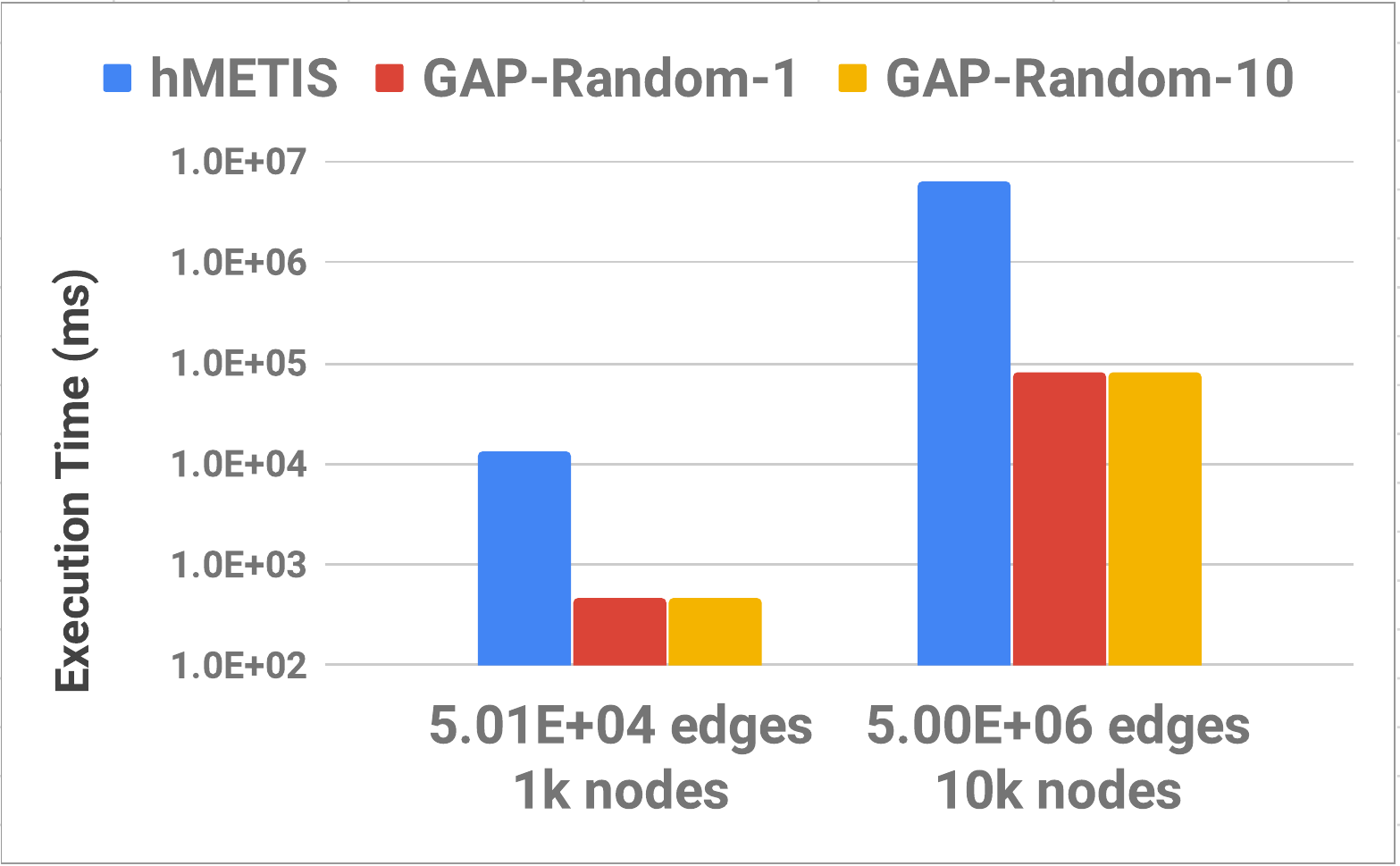}
        \caption{\small{The inference time of GAP is 10 to 100 times faster than hMETIS.}}
        \label{fig:random-exetime}
    \end{subfigure}
    \caption{Generalization of GAP on random graphs. GAP-Random-1 is trained on only one random graph, while GAP-Random-10 is trained on 10 random graphs. The result is the average over the 5 random graphs of 1k and 10k nodes. Performance of GAP-Random-1 and GAP-Random-10 is almost the same as hMetis but the inference time is 10 to 100 times faster than the runtime of hMETIS.}
    \label{fig:random}
\end{figure*}

\noindent\textbf{Baseline:} Since graph partitioning is NP-complete, solutions are generally derived using heuristics and approximation algorithms. While there has been a substantial amount of work on graph partitioning for specific graph structure/applications~\cite{gonzalez2012powergraph, hada2018}, \emph{hMETIS}~\cite{karypis_2000, karypis_1999} is a general framework that works across a wide variety of graphs and is shown to provide high quality partitions in different domains (e.g., VLSI, road network~\cite{miettinen_2006, xu2012hmetis}. Similar to hMETIS, GAP is a general framework that makes no assumptions about graph structure. In our experiments, we compare GAP against hMETIS. We set the hMETIS parameters to return balanced partitions with minimum edge cut.

\noindent\textbf{Performance Measures:} As we discussed in Section~\ref{sec:probDef}, balanced partitions with minimum edge cut is the goal of graph partitioning. We evaluate the performance of the resulting partitions by examining 1) \emph{Edge cut}: the ratio of the cut to the total number of edges, and 2) \emph{Balancedness}: is one minus the MSE of number of nodes in every partition and balances partition ($\frac{n}{g}$).

\subsection{Performance}
\label{sec:perf}
In this set of experiments, we find that GAP outperforms hMETIS. Since hMETIS does not generalize to unseen graphs and optimizes one graph at a time, we also constrain GAP to optimize one graph at a time for a fair comparison. We discuss the generalization ability of GAP in Section~\ref{sec:gen}. 

Table~\ref{tbl:perf} shows the performance of GAP against hMETIS on a 3-partition problem over real TensorFlow graphs. Both techniques generate very balanced partitions, with GAP outperforming hMETIS on edge cut for the VGG graph.

Figure~\ref{fig:edge_cut_random} shows the performance of GAP against hMETIS on random graphs when the number of partitions is varied from 2 to 10. The plots represent the average value across 5 random graphs. Both GAP and hMETIS produce 99\% balanced partitions. However, GAP is also able to find lower edge cut partitions than hMETIS. By examining the degree histograms of our datasets (Figures~\ref{fig:mnist-dist} to \ref{fig:random-1000-dist}), we found that while hMETIS heuristics work reasonably well on sparse TensorFlow graphs, GAP outperforms hMETIS on dense graphs.

\subsection{Generalization}
\label{sec:gen}
In this section, we show that GAP generalizes effectively on real and synthetic datasets. To the best of our knowledge, we are the first to propose a learning approach for graph partitioning that can generalize to unseen graphs.

\subsubsection{Generalization on real graphs}
In this set of experiments, we train GAP with a single TensorFlow graph, \textit{VGG}, and validate on \textit{MNIST-conv}. At inference time, we test the trained model on unseen TensorFlow graphs: \textit{AlexNet}, \textit{ResNet}, and \textit{Inception-v3}. 

Table~\ref{tbl:gen} shows the result of our experiments, and illustrates the importance of node features and graph embeddings in generalization. In \textit{GAP-id}, we use the index of a node as its feature, while in \textit{GAP-op}, the operation type (such as Add, Conv2d, and L2loss in TensorFlow) is used as the node feature. We encode all features as one-hots. Following Section~\ref{sec:framework}, we leverage Graph Convolution Networks~\cite{kipf2017semi} (GCN) and GraphSAGE~\cite{HamiltonYL17} to capture similarities across graphs. In \textit{GCN offline} and \textit{GraphSAGE offline}, we do not train the graph embedding module (Figure~\ref{fig:framework}) without gradient flow from the partitioning module, while in \textit{GraphSAGE trained} both modules are trained jointly. 
Table~\ref{tbl:gen} shows that \textit{GAP-op} with \textit{GraphSAGE trained} (last row) achieves the best performance and generalizes better than the other models. Note that this model is trained on a single graph, \textit{VGG} with only $1325$ nodes, and it is tested on \textit{AlexNet}, \textit{ResNet}, and \textit{Inception-v3} with $798$, $20586$, and $27114$ nodes, respectively. 

Figure~\ref{fig:inception_gen_vis} shows the GAP partitioning of \textit{Inception-v3} using a model trained on the same graph (\ref{fig:inception}) and a model trained on \textit{VGG} (\ref{fig:inception-gen}). Note that partitions are denoted by colors and we only show nodes whose operation type is convolution. In the scenario (\ref{fig:inception}) where we train and test GAP on \textit{Inception-v3}, we achieve 99\% balanced partitions with 4\% edge cut (Table~\ref{tbl:perf}). Where GAP is trained on \textit{VGG} and tested over the unseen graph (\textit{Inception-v3}), it achieves 99\% balanced partitions with 6\% edge cut (last row of Table~\ref{tbl:gen}). The partition assignments in Figures~\ref{fig:inception} and \ref{fig:inception-gen} are quite similar (75\%), which demonstrates GAP generalization.

We also observed that the similarity of the node features (operation types) in \textit{VGG} and other computation graphs used in inference and validation is correlated with the edge cut score of GAP partitioning (Figure~\ref{fig:similarity_op}). 
For example, let A and B be the set of the operation types in \textit{VGG} and \textit{ResNet}, respectively, with a Jaccard similarity of $\frac{|A \cap B|}{|A \cup B|} = 0.592$). Figure~\ref{fig:similarity_op} shows that as Jaccard similarity of a graph with \textit{VGG} increases, the edge cut decreases. In other words, the presence of similar node types across train and test graphs aids the generalization of our model.

\noindent\textbf{Model Architecture and Hyper-parameters:}
Here, we describe the details of the model with the best performance (corresponding to the last row of Table~\ref{tbl:gen}). The number of features (TensorFlow operation types) is 1518. GraphSAGE has 5 layers of 512 units with shared pooling, and the graph partitioning module is a 3 layer dense network of 64 units with a final softmax layer. We use ReLU as activation function and all weights are initialized using Xavier initialization~\cite{Xavier_2010}. We use the Adam optimizer with a learning rate of 7.5e-5.

\subsubsection{Generalization on synthetic graphs}
We further evaluate the generalization of GAP on \textit{Random} and \textit{Scale-free} graphs. Note that we train and test GAP on the same type of graph, but number of nodes may vary. For example, we train GAP on random graphs of 1k nodes and test on random graphs of 1k and 10k nodes. Similarly, we train GAP on scale-free graphs of 1k nodes and test on scale-free graphs of 1k and 10k nodes.

Figures~\ref{fig:scalefree-edgecut}, \ref{fig:scalefree-bal}, and \ref{fig:scalefree-exetime} show the edge cut, balancedness, and execution time of GAP against hMETIS over the scale-free graphs (every point is the average of 5 experiments). 
In GAP-Scalefree-1 we train GAP with only one scale-free graph, while GAP-Scalefree-10 is trained on 10 scale-free graphs. We then test the trained models GAP-Scalefree-1 and GAP-Scalefree-10 over 5 unseen scale-free graphs of 1k and 10k nodes and we report the average results. Figure~\ref{fig:scalefree-edgecut} shows that both GAP-Scalefree-1 and GAP-Scalefree-10 partition the unseen graphs of 1k and 10k nodes with lower edge cut than hMETIS. Despite the balancedness of GAP-Scalefree-1 being lower than that of hMETIS, by increasing the number of graphs in the training set (GAP-Scalefree-10) balancedness is improved as shown in Figure \ref{fig:scalefree-bal}, while its edge cut is still smaller (\ref{fig:scalefree-edgecut}). Furthermore, GAP-Scalefree-10 runs slightly faster than hMETIS (\ref{fig:scalefree-exetime}) and its partitions are just as balanced as those of hMETIS (\ref{fig:scalefree-bal}) but with lower edge cut (\ref{fig:scalefree-edgecut}).


Figures~\ref{fig:random-edgecut}, \ref{fig:random-bal}, and \ref{fig:random-exetime} show the edge cut, balancedness, and execution time of GAP against hMETIS on random graphs. Every point is the average of 5 experiments. 
In GAP-Random-1, we train GAP on only one random graph, while in GAP-Random-10, we train on 10 random graphs. We then test the trained models GAP-random-1 and GAP-Random-10 on 5 unseen random graphs with 1k and 10k nodes and we report the average results. The performance of GAP when generalizing on unseen random graphs of 1k and 10k nodes is almost the same as the performance of hMETIS, while Figure~\ref{fig:random-exetime} shows that during inference, GAP is 10 to 100 times faster than the runtime of hMETIS.

\noindent\textbf{Model Architectures and Hyper-parameters:}
Unlike computation graphs where node features are operation types, nodes in synthetic graphs have no features. Furthermore, we must train a model that generalizes to graphs of different sizes. For example, we train a model on a random graph with 1k nodes and test it on a random graph with 10k nodes. To do so, we apply PCA to the adjacency matrix of a featureless graph and retrieve embeddings of size 1000 as our node features. We use ReLU as our activation function and all weights are initialized using Xavier initialization. We also use the Adam optimizer. Here are the rest of the hyperparameters for each model.

    \noindent\emph{GAP-Scalefree-1:} model is trained with one scale-free graph. GraphSAGE has 5 layers of 512 units, and graph partitioning module is 3 layer dense network of 128 units with softmax. Learning rate is 2.5e-6.
    
    \noindent\emph{GAP-Scalefree-10:} Trained with 10 scale-free graphs. GraphSAGE has 4 layers of 128 units, and graph partitioning module is 1 layer dense network of 64 units with softmax. Learning rate is 7.5e-6.
    
    \noindent\emph{GAP-Random-1:} Trained with only random graph. GraphSAGE has 5 layers of 128 units with shared pooling, and graph partitioning module is 2 layer dense network of 64 units with softmax. Learning rate is 7.5e-4.
    
    \noindent\emph{GAP-Random-10:} Trained with 10 random graphs. GraphSAGE has 2 layers of 256 units with shared pooling, and graph partitioning module is 3 layer dense network of 128 units with softmax. Learning rate is 7.5e-6.

\vspace{-0.1in}
\section{Conclusion}
\label{sec:conc}
We propose a deep learning framework, GAP, for the graph partitioning problem, where the objective is to assign the nodes of a graph into balanced partitions while minimizing the edge cut across the partitions. Our GAP framework enables generalization: we can train models that produce performant partitions at inference time, even on unseen graphs. This generalization is an advantage over existing baselines which redo the optimization for each new graph. Our results over widely used machine learning models (ResNet, VGG, and Inception-v3), scale-free graphs, and random graphs confirm that GAP achieves competitive partitions while being up to 100 times faster than the baseline and generalizing to unseen graphs.
\bibliography{gp}

%
%

\bibliographystyle{apalike}

\end{document}